\pdfoutput=1

\documentclass[11pt]{article}

\usepackage[]{EMNLP2023}

\usepackage{times}
\usepackage{latexsym}

\usepackage[T1]{fontenc}

\usepackage[utf8]{inputenc}

\usepackage{microtype}

\usepackage{inconsolata}

\usepackage{graphicx}

\title{LOKE: Linked Open Knowledge Extraction for Automated Knowledge Graph Construction}

\author{Jamie McCusker \\
Rensselaer Polytechnic Institute \\
Troy, NY 12180, USA \\
\texttt{mccusj2@rpi.edu}}

\begin{document}
\maketitle
\begin{abstract}
While the potential of Open Information Extraction (Open IE) for Knowledge Graph Construction (KGC) may seem promising, we find that the alignment of Open IE extraction results with existing knowledge graphs to be inadequate.
The advent of Large Language Models (LLMs), especially the commercially available OpenAI models, have reset expectations for what is possible with deep learning models and have created a new field called prompt engineering. 
We investigate the use of GPT models and prompt engineering for knowledge graph construction with the Wikidata knowledge graph to address a similar problem to Open IE, which we call Open Knowledge Extraction (OKE) using an approach we call the Linked Open Knowledge Extractor (LOKE, pronounced like ``Loki'').
We consider the entity linking task essential to construction of real world knowledge graphs.
We merge the CaRB benchmark scoring approach with data from the TekGen dataset  for the LOKE task.
We then show that a well engineered prompt, paired with a naive entity linking approach (which we call LOKE-GPT), outperforms AllenAI's OpenIE 4 implementation on the OKE task, although it over-generates triples compared to the reference set due to overall triple scarcity in the TekGen set.
Through an analysis of entity linkability in the CaRB dataset, as well as outputs from OpenIE 4 and LOKE-GPT, we see that LOKE-GPT and the ``silver'' TekGen triples show that the task is significantly different in content from OIE, if not structure.
Through this analysis and a qualitative analysis of sentence extractions via all methods, we found that LOKE-GPT extractions are of high utility for the KGC task and suitable for use in semi-automated extraction settings.

\end{abstract}

\section{Introduction}

The introduction of the most recent generation of Large Language Models (LLMs), especially Generative Pre-trained Transformer (GPT) 3.5 and 4 models, has shown significant capability in a wide range of tasks around continuation of fluent, plausible text \citep{openai2023gpt4}.
GPT 3.5 and later has also shown significant capabilities when used in prompt engineering, where a significant amount of instruction and data input is supplied to the model as a prompt to continue.
Called prompt engineering, this approach is similar to programming, except the goal is to place the model in a very specific activation space that can lead to useful outputs.

Our initial investigations into prompt engineering extend the GraphGPT\footnote{https://github.com/varunshenoy/GraphGPT} prompt, which is a toy prompt used by users to describe graphs to a visual charting tool. 
Such prompts are able to perform simultaneous entity and relation extraction by producing JSON-based lists of \emph{subject, predicate, object} triples.
With further prompt engineering, we found we were able to produce qualitatively useful and relevant graphs that could easily be linked to Wikidata entities and properties to produce usable Resource Description Framework (RDF) \citep{manola2004rdf} knowledge graph fragments.
An important aspect of this approach is that there is no pre-training, and the predicates generated by the model are entirely open-ended.
We often found useful properties that were missing from the Wikidata \citep{vrandevcic2014wikidata} property vocabulary, while existing relationship extraction models are usually limited to a fixed vocabulary.
Additionally, we found that this approach was very successful at extracting multiple entities and relationships from single sentences that would require detailed annotation of relationships from text.

In this paper we found that engineered prompts using the OpenAI text-davinci-003 model provide potentially competitive performance to task-specific KG construction from text, and that existing benchmarks and OpenIE algorithms are not capable of producing graphs suitable for creation of real world knowledge graphs.

\section{Background}

Within the Semantic Web and knowledge representation fields, we define a knowledge graph as a form of directed labeled graph $G=\{V,E\}$ where each edge $e \in E$  is a statement, or triple, of the form $e = (subject, predicate, object)$, where $subject,object \in V$, and $subject \in S$, $predicate \in P$, and $object \in O$. The $subject$ is the source vertex, the $predicate$ is the edge label, and the $object$ is the target vertex. 

Within Resource Description Framework (RDF)-based knowledge graphs \citep{manola2004rdf}, the following rules apply:
\begin{enumerate}
    \item  $\forall v \in V \Rightarrow v \in R \cup D$ where $R=I \cup U$ is the set of all entities, or resources, that are either identified by a Uniform Resource Identifier, or URI, ($v\in I$) or unidentified ($v\in U$); and $D$ is the set of all literal data values, aka strings, numbers, and other data values.
    \item $S \subseteq R$: all subjects are resources.
    \item $P \subseteq I$: all predicates are identified resources. Each resource used as a predicate is called a \emph{property}.
    \item $O \subseteq R \cup D$: all objects can be resources or literals.
    \item $\forall v \in I$, any use of $v$ is universally unambiguous: all uses of $v$ always denote the same entity regardless of the knowledge graph it belongs to.
\end{enumerate}

As a consequence of Rule 5, it is trivial to integrate knowledge graphs that use the same URIs simply by taking the union of the two graphs.
Within this paradigm, there is a theoretical ``universal knowledge graph'' that consists of the union of all expressed $G$s in the universe.
Steps have been taken towards making this universal graph more accessible through the conventions around Linked Open Data (LOD), \citep{bizer2011linked} by focusing on the use of HTTP URLs for identifiers and providing statements about the entity denoted by that URL when requested.

Consequently, entity and property linking is a crucial step to accurate knowledge graph construction from text.
Subjects and objects are considered \emph{symmetric}, in that the same level of detail should be included in each, so that if an entity appears as a subject or object, it will be linked to the correct URI within the knowledge graph.

While Wikidata includes reification of these edges to allow for further annotation, and other knowledge graphs allow for subgraph labeling using named graphs, all knowledge graphs using RDF conform or map to these semantics, and many (including Wikidata) provide LOD and are therefore part of this ``universal knowledge graph.''

\section{Algorithm Approach}

The LOKE-GPT processor is based on the OpenAI text-davinci-003 model using the completion API. 
We started with the stateless prompt\footnote{\url{https://github.com/varunshenoy/GraphGPT/blob/main/public/prompts/stateless.prompt}} used in the GraphGPT project, which allows users to construct small graphs using conversational prompts.
The prompt was modified (see Appendix \ref{sec:prompt_appendix}) to refine the kind of entities extracted and the properties used to link them.
The prompt reliably generates a JSON list of lists, of the form \texttt{[["subject", "predicate", "object", "data type"], ... ]}.
The data type is for literal values, such as dates and quantities.
It is missing for links between entities.
We ignore the object type for this evaluation, but it is useful for constructing RDF with higher quality literal values.

We employ Wikidata-based entity linking for both RDF construction and to improve triple quality for evaluation.
We use a simple entity linking algorithm based partial matches against a full text index, and then take the entity with the match with the smallest edit distance.
In the case of Wikidata entities, we use the index from the spaCy entity linker project\footnote{https://github.com/egerber/spacy-entity-linker}, which provides an entity index based on the \citet{kensho-derived-wikimedia-data} entity dump from Wikidata.
This dump was published in 2020, so is an incomplete index of current Wikidata entities.
We also created a text search index of the Wikidata properties using the Whoosh full text search library. 
Indices are searched for entities and properties, but do not take advantage of context. 
We then rank the top ten hits by their edit distance from the search term.
We compute a confidence score for the match from the edit distance, by treating each edit as an uncertain probability of success (\(p=0.999\)), where the joint probabilities over edit distance ($\varepsilon$) form the following exponential decay to full uncertainty (\(u=0.5\)) for confidence $c$:

\[c = (1 - u) + u p^\varepsilon\]

\begin{figure}
    \centering
    \includegraphics[width=\linewidth]{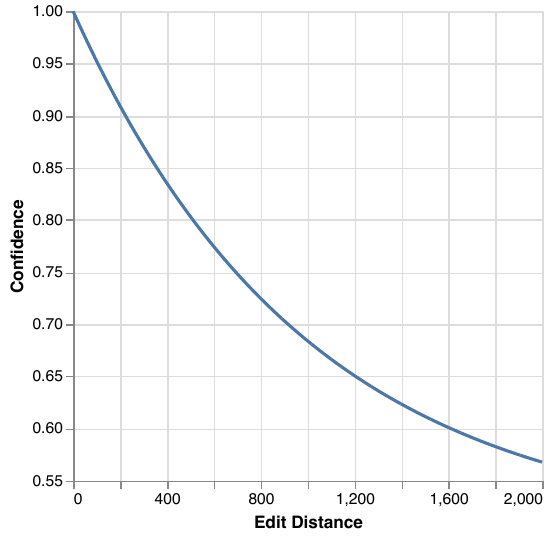}
    \caption{The function computing link confidence as a result of the edit distance from the query label and the label hit, where \(p=0.999\) and \(u=0.5\).}
    \label{fig:edit_distance_confidence}
\end{figure}

This results in the curve shown in Figure \ref{fig:edit_distance_confidence}.
To compute statement confidence, we then take the joint confidence of the confidence for the subject, predicate, and object for object statements, or the subject and predicate for literal statements.

Finally, for evaluation in the TekGen dataset, we correct the labels for each linked entity and property based on the preferred label for each, normalizing any aliases or partial matches. 

\section{Evaluation Methods}

In \citet{bhardwaj-etal-2019-carb}, the use of  CaRB  shows that Open IE 4 provides the best performance on that benchmark, so we evaluate LOKE-GPT against Open IE 4 from \citet{stanovsky-etal-2018-supervised} on the OKE task.
1000 sentences were randomly selected from the TekGen validation dataset, and then filtered so that sentences mention the subjects from the extracted statements, leaving 607 sentences.
For each sentence, statements are extracted from both LOKE-GPT both with and without linked entity correction, as well as  from Open IE 4. 
The optimal and curved Precision ($P=\frac{tp}{tp+fp}$), Recall ($R=\frac{tp}{tp+fn}$), and F-1 ($F_1=2\cdot\frac{P \cdot R}{{P + R}}$) scores are then computed using the CaRB scoring algorithm using the most lenient triple matching approach.
Additionally, we evaluate the CaRB gold dataset, TekGen's statements, and OpenIE 4 and LOKE-GPT outputs of the TekGen validation subset for it suitability as contributions to a Wikidata-like knowledge graph by reporting the fraction of subjects, predicates, objects, and whole triples that are fully linked to entities and properties in the Wikidata knowledge graph.

We used a subset of the TekGen Dataset from \cite{agarwal-etal-2021-tekgen} for testing and validation. 
TekGen was prepared for generation of text from knowledge graph fragments, and contains knowledge graph assertions from Wikidata that they found were expressed in associated sentences from Wikipedia articles. 
The assertions are written using the human readable labels of URIs instead of the URIs themselves, which makes them good candidates for LLM inputs.
However, we had to re-link them for evaluation. 
Because of the cost and rate limiting with OpenAI models, we prepared a random sample of sentences from validation (1000 sentences, filtered for subject mentions in the text to 607 sentences) sets to process using the engineered prompt. 

\section{Results}

We evaluate the suitability of OpenIE 4 and LOKE-GPT for the OKE task through a qualitative assessment of statement extraction quality and relevance, the performance of OpenIE 4 and LOKE-GPT against the CaRB-scored TekGen validation sample, and an evaluation of the ability of both to link extracted entities and properties in the same data, as well as a comparison against the linkability of the TekGen reference statements and the CaRB gold statements.
We find that in all cases LOKE-GPT outperforms OpenIE 4, and, while it does not exactly extract triples that correspond to the ones available in TekGen, the extracted triples are reflective of the source text and generally representable using the entities and properties available in Wikidata.

\subsection{Qualitative Assessment}

We use the following example sentences from the TekGen dataset to assess the quality of knowledge extraction:
\begin{enumerate}
    \item Tiram is a town and Village Development Committee in Pyuthan, a Middle Hills district of Rapti Zone, western Nepal.
    \item Toxabramis maensis is a species of ray-finned fish in the genus Toxabramis.
    \item Bahaa al-Farra (born 10 March 1991) is a Palestinian runner from Gaza.
\end{enumerate}

These are the triples from Wikidata that are corroborated by the sentence:

\begin{enumerate}
    \item Tiram; country; Nepal
    \item Toxabramis maensis; taxon rank, Species
    \item Bahaa al-Farra; date of birth; 10 March 1991
\end{enumerate}

It is notable is that the available triples are not at all reflective of all the knowledge encoded in the sentences.
However, the actual entities in the statements are easy to link to Wikidata entities (since they were derived from them).

OpenIE 4 produces the following statements for those sentences:
\begin{enumerate}
    \item Tiram; is; a town and Village Development Committee in Pyuthan
    \item Toxabramis maensis; is; a species of ray-finned fish in the genus
        \item Bahaa al-Farra; is; a Palestinian runner from Gaza
\end{enumerate}

This extraction has two issues: one, the statements extracted only reflect the main clause of  the sentences, and the linkability of the extracted objects is much lower. 
None of the object entities are ones that can be found in Wikidata, as they are too specific and would generally be decomposed into finer grained, more generalized entities.
Additionally, knowledge graphs expect representational symmetry between subjects and objects.
The objects extracted by OpenIE 4 (and the ones tagged in the CaRB dataset) are more likely to be compound nouns, as well as a mix of descriptors, types, and referents.
As a result, the complex noun phrases that are extracted are not good candidates for representation as singular entities as modeled in most knowledge graphs.
This representational asymmetry means that the object of one statement extracted from OpenIE 4 and other Open IE models is highly unlikely to be used as the subject of another statement. 
This results in disconnected stars of statements, instead of the interconnected network of entities that knowledge graphs generally try to build.

Conversely, LOKE-GPT produces these statements:
\begin{enumerate}
    \item Tiram; type; town \\
    Tiram; type; Village Development Committee \\
    Tiram; location; Pyuthan \\
    Pyuthan; type; Middle Hills district  \\
    Pyuthan; location; Rapti Zone \\
    Rapti Zone; location; western Nepal
    \item Toxabramis maensis; species; ray-finned fish \\
    Toxabramis maensis; genus; Toxabramis
    \item Bahaa al-Farra; born; 10 March 1991 \\
    Bahaa al-Farra; citizenship; Palestine \\
    Bahaa al-Farra; occupation; runner \\
    Bahaa al-Farra; location; Gaza
\end{enumerate}

What is immediately apparent is the increase in the number and quality of triples, as well as the representational symmetry of the subjects and objects.
For instance, nearly every single object in these statements is resolvable as an entity in Wikidata, whereas none of the objects extracted from OpenIE 4 have corresponding Wikidata entities.
There are still some issues with how taxonomy is expressed for Toxabramis maensis, but this can be assisted by introducing alternate labels for properties.
It is worth noting that the modeling of species in Wikidata is fairly specialized and not easily aligned with the way it is phrased here.
While ``Bahaa al-Farra; born; 10 March 1991'' is the only close triple found in the TekGen dataset, all the facts about Tiram, Pyuthan, and Rapti Zone are confirmed by Wikidata. 
Additionally, the following relevant facts are listed on the entry for Bahaa al-Farra: ``date of birth; 10 March 1991'', ``country of citizenship; State of Palestine'', and ``occupation; athletics competitor''.

\subsection{TekGen-CaRB Benchmark}

\begin{figure}[!ht]
    \centering
    \includegraphics[width=\linewidth]{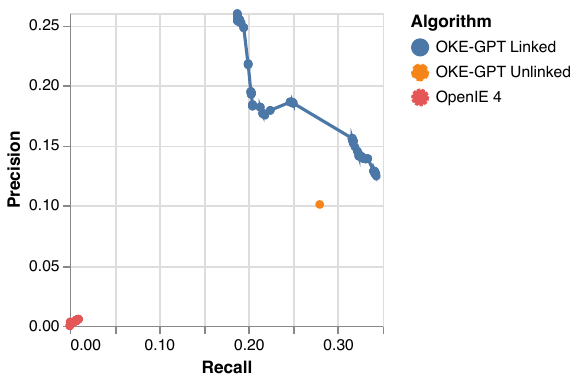}
    \caption{The precision/recall curves for OpenIE 4, link-corrected OKE-GPT (OKE-GPT L), and uncorrected OKE-GPT output (OKE-GPT U). OKE-GPT U does not have confidence levels available, so there is only one point on the curve. The OpenIE 4 scores are sufficiently low that it is difficult to discern as a curve.}
    \label{fig:prf1}
\end{figure}

Results for the TekGen-based benchmark using CaRB metrics show a clear advantage on Wikidata-style statements from LOKE-GPT, even on incomplete datasets like TekGen. Table \ref{tab:prf1} shows a 31-fold improvement in optimized F1 score of link-corrected LOKE-GPT over the prior best performer, OpenIE 4. Additionally, the precision/recall curve using link confidence with corrected and uncorrected LOKE-GPT shows marked improvement in all configurations against OpenIE 4 in Figure \ref{fig:prf1}. 

\begin{table}[!ht]
\begin{tabular}{|l|c|c|c|c|}
\hline
Algorithm	& AUC	& P	& R	& F1 \\ \hline
OpenIE 4	& 0.000	& 0.005	& 0.009	& 0.007 \\
OKE-GPT L	& 0.145	& \textbf{0.248}	& \textbf{0.195}	& \textbf{0.218} \\
OKE-GPT U	& \textbf{0.154}	& 0.101	& 0.28	& 0.148 \\
\hline
\end{tabular}
\caption{The optimized performance of OpenIE 4, link-corrected OKE-GPT (OKE-GPT L), and uncorrected OKE-GPT output (OKE-GPT U). For each, Area under curve (AUC), precision (P), recall (R), and F-1 score (F1) are reported. }
\label{tab:prf1}
\end{table}

\subsection{Entity Linkability}
\begin{figure}[!ht]
    \centering
    \includegraphics[width=1\linewidth]{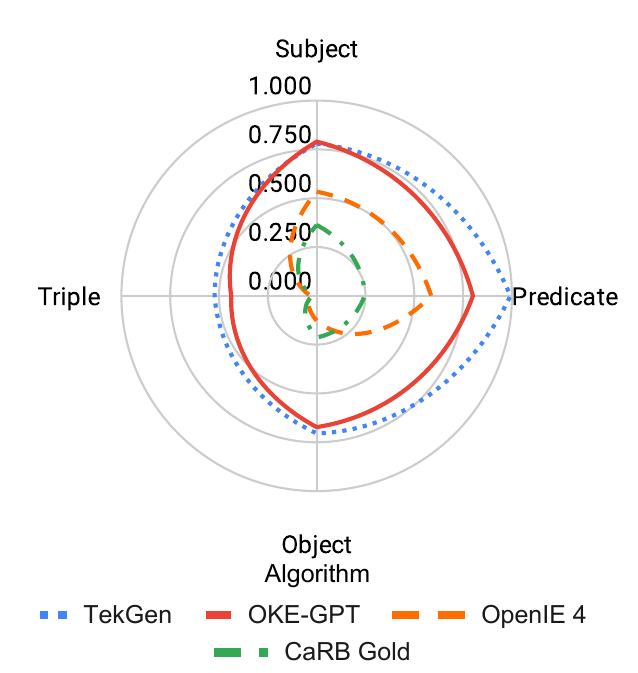}
    \caption{A radar plot of the fraction of subjects, predicates, objects, and full triples in two approaches (OKE-GPT and OpenIE 4) and two benchmarks (TekGen and CaRB Gold). The lower linkability of OpenIE 4 and CaRB show their lack of suitability to represent entities and properties from real world knowledge graphs like Wikidata.}
    \label{fig:linkability}
\end{figure}

As shown in Table \ref{tab:linkability} and Figure \ref{fig:linkability}, subjects, objects, and predicates from OKE-GPT are far more likely to be linkable to Wikidata entities and properties.
OKE-GPT follows a sufficiently similar representational approach that the linkability numbers are highly congruent between OKE-GPT and TekGen, even though OKE-GPT was is a one-shot prompt that was developed before examining the TekGen dataset.
Conversely, OpenIE 4 has much lower performance across the board, and is clearly trying to solve the Open IE problem, which does not focus on subject/object symmetry (as shown in the linkability performance of OpenIE 4 and CaRB).
While objects are less linkable in general across datasets, in the case of TekGen and OKE-GPT, this seems to be at least partly the result of including literal values as objects in statements, which are not inherently linkable.
Figure \ref{fig:linkability} in particular highlights the performance similarity of OKE-GPT and TekGen, and then OpenIE 4 and Carb Gold.
\begin{table}[!ht]
\begin{tabular}{|l|c|c|c|c|}
\hline
Algorithm	& S	& P	& O & T \\
\hline
TekGen	& 0.774	& 0.997	& 0.700	& 0.525 \\
OKE-GPT	& \textbf{0.793}	& \textbf{0.802}	& \textbf{0.670}	& \textbf{0.437} \\
OpenIE 4	& 0.532	& 0.587	& 0.123	& 0.031 \\
CaRB	& 0.364	& 0.247	& 0.211	& 0.022 \\
\hline
\end{tabular}
\caption{The fraction of subjects ($S$), predicates ($P$), objects ($O$), and full triples ($E$) that are linkable within the TekGen validation sample, OKE-GPT,  OpenIE 4, and the CaRB gold test dataset. }
\label{tab:linkability}
\end{table}

\section{Discussion}

The symmetry of entities in knowledge graphs is an important component to their success.
While there are often larger scale descriptions that are included in the graph, entities must be general purpose to facilitate reuse.
One challenge with asymmetric statements like the ones extracted by OpenIE 4 is that, when an entity is new to the knowledge graph, it becomes difficult to reuse any kind of generated entities in other statements.
While it is probably possible to link the complex entities that are used in Open IE 4, they are often ambiguous and may result in lost information.
Additionally, most real world knowledge graphs like Wikidata are built from multiple sources. Wikidata is an agglomeration of knowledge from existing databases, as well as from human contributions.
As a result, the OKE task needs to be reflective of the sorts of entities that are curated by humans manually and from real world data.

It is also interesting that the text-davinci-003 model is so instructable in this task. 
By providing a set of instructions and corresponding exemplars, it becomes reasonably successful in tasks that have required large scale specialized training or detailed heuristic extraction approaches to perform.
An interesting aspect for this is that, since the model does no explicit sentence parsing, it seems to be better able to re-compose entities that might be difficult to tease out of a sentence if the entities were simply quoted from the sentence.
For instance, while, at a conceptual level, the statement ``Bahaa al-Farra; citizenship; Palestine'' exists in the sentence ``Bahaa al-Farra (born 10 March 1991) is a Palestinian runner from Gaza'', it is only explicitly stated as ``Bahaa al-Farra... is a Palestinian''. The re-composition of Palestinian to Palestine and use of the predicate ``citezenship'' was automatically incorporated into the graph.
This, by itself, is extremely useful in that it is able to recognize entities and relations that are not explicitly named in the sentence.

\subsection{Future Work}
This is still preliminary work and can be expanded in many directions.
First, we hope to improve the approach to entity and relation linking beyond simple search to an context-aware embedding-based search.
The use of OpenAI's models limits our ability to scale usage and evaluation of this approach. 
At the time of our experiments, found that each sentence extracted through the OpenAI API cost approximately \$0.01 USD, after inclusion of the necessary prompts.
We will be investigating the use of the Alpaca LLM from \citet{alpaca} as a local replacement for text-davinci-003.
Further, we hope that we can develop an approach to train smaller language models using semi-supervised data distillation to perform specialized knowledge extraction.
Additionally,  it is clear from the literature and our own experiments that the utility of knowledge graphs constructed using these approaches is at least as important as triple matching, so we hope to undertake functional evaluations of large scale knowledge graphs generated by our approach.
We hope to also evaluate these approaches by generating domain specific knowledge graphs with customized engineered prompts, and assess the multi-lingual capabilities of such models.

\section{Related Work}
This is the first paper to our knowledge to assess the performance of open knowledge extraction using one-shot learning with large language models.
The field of knowledge graph construction from text has been very active, however.
Open IE approaches (such as \citet{jaradeh2023information} and \citet{muhammad2020open}), have been evaluated, but most of these seem to be evaluated on consensus extraction datasets like \citet{elsahar-etal-2018-rex}, leading to the challenge that approaches are only evaluated against existing baselines and not a complete picture of what is actually being expressed in the supplied text.
As explored in \citet{li2019survey}, approaches for entity recognition and separate relation extraction have been proposed, however these approaches seem to generally assume that only one relationship between entities is available, and often use closed sets of relations that do not allow for expansion beyond the trained relations.

There is a large body of research that uses non-Open IE approaches for knowledge extraction.

Often knowledge extraction and to our knowledge, none 

\section*{Limitations}
Currently, we have only been successful at prompt engineering against the text-davinci-003 model from OpenAI, restricting the scalability to availability of OpenAI resources and and research budget. 
Attempts to reproduce this approach using the LLAMA models have failed, as the models cannot produce relevant JSON, and the ability to produce valid JSON structures is also impaired.
We will also note that we have only attempted naive entity linking approaches.
Entity linking using embeddings and other non-exact indices may provide increased accuracy.
As the benchmarks  \citep{agarwal-etal-2021-tekgen,bhardwaj-etal-2019-carb} in this round were English only, we have only been able to evaluate this against English text.
Due to API availability, we were unable to perform a full analysis of all the validation data available in TekGen.
Finally, due to API availability, we have not performed an ablation study of the prompt engineering that has gone into the current work, nor have we attempted to optimize the examples given in the prompt.

\section{Conclusion}

We show that the Open Knowledge Extraction (OKE) task is significantly different from the Open Information Extraction (Open IE) task, and that existing approaches to Open IE, while similar structurally, are unsuited to solving OKE.
We also show that an engineered prompt used with the text-davinci-003 model can effectively extract fine grained, symmetric entities as subjects and objects from plain text that are suitable for resolution against existing real-world knowledge graphs like Wikidata. 
We also show that the accuracy of such extraction is improved through entity and property link correction, and that comparisons of the output between this approach, OKE-GPT, and the existing best Open IE approach, OpenIE 4, show significant performance benefits to OKE-GPT when compared using the CaRB evaluation algorithm and a sample of validation data from the TekGen dataset.
Finally, we hope to further this research by performing functional evaluations of domain-specific corpora and to abstract this capability away from any one LLM for more generally available use, and feel that the output from such algorithms is suitable for knowledge graph construction in general purpose cases.

\section*{Ethics Statement}
As with all knowledge extraction tools, it is possible to use tools like this to aggregate knowledge about entities that are attempting, for legitimate reasons, to obscure that knowledge about themselves.
While on the one hand, knowledge about scientific domains becomes much easier to query, as does information gleaned about potential criminals, it is also possible to learn more about private citizens who otherwise wish to remain anonymous.
It is also possible to use tools like this to track the activities and other aspects of political dissidents in oppressive regimes.
Because of the nature of open knowledge extraction, it becomes difficult to tease apart the ability to automatically aggregate knowledge in these domains, and it is futile to rely on the incompetence of evil.

\section*{Acknowledgements}

This research was supported 

\bibliography{custom}
\bibliographystyle{acl_natbib}

\appendix

\section{Appendix: KGIE Engineered Prompt}
\label{sec:prompt_appendix}

The GraphGPT prompt provides the basis for our prompt, and provides some basic capabilities that we build off of:

\begin{verbatim}
Given a prompt, extrapolate as many 
relationships as possible from it and 
provide a list of updates.

If an update is a relationship, provide 
[ENTITY 1, RELATIONSHIP, ENTITY 2]. The 
relationship is directed, so the order 
matters.

If an update is related to a color, provide 
[ENTITY, COLOR]. Color is in hex format.

If an update is related to deleting an 
entity, provide ["DELETE", ENTITY].

Example:
prompt: Alice is Bob's roommate. 
Make her node green.
updates:
[
 ["Alice", "roommate", "Bob"], 
 ["Alice", "#00FF00"]
]

prompt: $prompt
updates:
\end{verbatim}

We do not need to remove entities or give a color for an entity, so we simplified the prompt to remove those functions.
Upon evaluation against example sentences, we also found that the implicit granularity of entities was insufficient, so we included instructions to improve that granularity.
We also instructed the model to use Wikidata and similiar links between entities, and to provide types for literals when they are expressed.
The final prompt that we used for the evaluation is as follows:

\begin{verbatim}
Given a prompt, extrapolate as many 
relationships as possible from it and 
provide a list of many fine-grained 
simple links.

If a link is a relationship, provide 
[ENTITY 1, RELATIONSHIP, ENTITY 2].

Use modeling approaches as similar as 
possible to Wikidata.

Links must correspond to Wikidata 
properties.

Links must be the simplest possible 
relationships between as many entites 
as possible.

The relationship is directed, so the 
order matters.

If the link contains entities, it should 
be broken apart into multiple entities 
and relationships.

Entities should not contain 'and', but 
should be broken into the smallest 
possible groupings.

If the link is a relationship between an 
entity and a value, the values should have 
their data type after them, in the form 
[ENTITY 1, RELATIONSHIP, VALUE, TYPE].

Roles should be expressed as relationships 
between entities.


Example:

prompt: Alice is Bob's roommate, they live 
in New York City. Alice, an american 
business analyst in insurance, was born 
in 1983.

updates:
[
 ["Alice", "roommate", "Bob"], 
 ["Alice","location”, "New York City"],  
 ["Bob","location", "New York City"], 
 ["Alice", "born", "1983", "year"], 
 ["Alice", "citizenship", "America"], 
 ["Alice", "occupation", "business analyst"], 
 ["Alice", "domain", "insurance"]
]

prompt: $prompt

updates:
\end{verbatim}

\end{document}